\documentclass[11pt]{article}

\usepackage[preprint]{acl}

\usepackage{times}
\usepackage{latexsym}
\usepackage{booktabs}
\usepackage{multirow}
\usepackage{hyperref}
\usepackage{makecell}
\usepackage{listings}
\usepackage{xcolor}

\usepackage{graphicx}
\usepackage{array}
\usepackage[export]{adjustbox}
\UseRawInputEncoding
\usepackage[normalem]{ulem}

\newcommand{\contextmemeimgtext}[2]{%
\begin{minipage}[t]{\linewidth}
    \centering
    \adjustbox{valign=t, max width=0.65\linewidth, max height=1.6cm}{%
        \includegraphics{#1}%
    }%
    \par\vspace{0.15em}
    \raggedright
    \scriptsize #2
\end{minipage}
}

\newcommand{\ctxcellsep}{%
\vspace{0em}
\par\noindent\textcolor{gray!60}{\rule{\linewidth}{0.3pt}}
\vspace{0.25em}
}

\providecommand{\ctxcellsep}{%
\par\vspace{0.35em}
\noindent\textcolor{gray!60}{\rule{\linewidth}{0.3pt}}
\par\vspace{0.25em}
}

\providecommand{\contextblock}[1]{%
\begin{minipage}[t]{\linewidth}
\raggedright
#1
\end{minipage}
}

\providecommand{\contextmemeimgtext}[2]{%
\begin{minipage}[t]{\linewidth}
\centering
\includegraphics[width=0.95\linewidth]{#1}
\par\vspace{0.35em}
\raggedright
#2
\end{minipage}
}

\lstdefinestyle{promptstyle}{
  basicstyle=\ttfamily\scriptsize,
  breaklines=true,
  breakatwhitespace=true,
  breakautoindent=false,
  breakindent=0pt,
  columns=fullflexible,
  keepspaces=true,
  showstringspaces=false,
  frame=single,
  linewidth=\dimexpr\columnwidth-0.5em\relax,
  xleftmargin=1.5em,
  xrightmargin=0pt,
  framexleftmargin=0pt,
  framexrightmargin=0pt,
  aboveskip=0.5em,
  belowskip=0.5em
}

\usepackage{amsmath}

\usepackage[T1]{fontenc}
\usepackage[utf8]{inputenc}

\usepackage{xcolor}
\usepackage{CJKutf8}

\usepackage{soul}  % strikethrough (\st) for resolved comments
\usepackage{microtype}

\usepackage{inconsolata}

\usepackage{enumitem}

\usepackage{titletoc}

\usepackage{tikz}
\usetikzlibrary{positioning,calc,arrows.meta}

\usepackage{graphicx}
\graphicspath{{pic/}{latex/pic/}}
\usepackage{float} % enables [H] float placement option for figures/tables

\newcommand{\safeincludegraphics}[2][]{%
  \IfFileExists{#2}{%
    \includegraphics[#1]{#2}%
  }{%
    \fbox{%
      \parbox[c][0.22\textheight][c]{0.9\linewidth}{%
        \centering Missing figure file\\
        \texttt{\detokenize{#2}}%
      }%
    }%
  }%
}

\title{I Know What You Meme, Even If it Emerged Today: Understanding Evolving Memes through Open-World Knowledge Acquisition}
\author{
    Shanhong Liu\textsuperscript{\rm 1},
    Rui Cao\textsuperscript{\rm 1},
    Pai Chet Ng\textsuperscript{\rm 2},
    De Wen Soh\textsuperscript{\rm 1} \\
    \textsuperscript{\rm 1}Singapore University of Technology and Design \\
    \textsuperscript{\rm 2}Singapore Institute of Technology \\
    \texttt{shanhong\_liu@mymail.sutd.edu.sg},
    \texttt{\{rui\_cao,dewen\_soh\}@sutd.edu.sg}, \\
    \texttt{paichet.ng@singaporetech.edu.sg}
}

\begin{document}
\maketitle

\begin{abstract}
Multimodal memes are dynamic and often require up-to-date background knowledge for interpretation.
Existing methods often overlook such knowledge or rely on fixed parametric knowledge of pre-trained models that may be incomplete, outdated, or unavailable for emerging memes.
We introduce \textit{Query-Retrieve-Conclude}, a zero-shot framework that identifies missing knowledge, retrieves open-web evidence, and synthesizes evidence-grounded background knowledge for meme understanding and detection. We also introduce a curated meme understanding benchmark of recent memes from 2024–2026 with external background knowledge annotations.
Experiments on three meme understanding datasets and five meme detection tasks show that our framework improves knowledge recovery, meme understanding and downstream detection over zero-shot baselines.

\end{abstract}

\section{Introduction}
%\crcomment{Your previous version was in prev_intro.tex} 

Internet {multimodal} memes are inherently dynamic and temporally evolving forms of online communication.\footnote{\url{https://en.wikipedia.org/wiki/Memetics}} Their meanings often depend on recent news, viral templates, emerging public controversies, cultural references, or newly formed associations between public figures and events \cite{valensise2021entropy}. 
\begin{figure}[t!]
    \centering
    \includegraphics[
        width=\linewidth,
        trim={0cm 0cm 0cm 0cm},
        clip
    ]{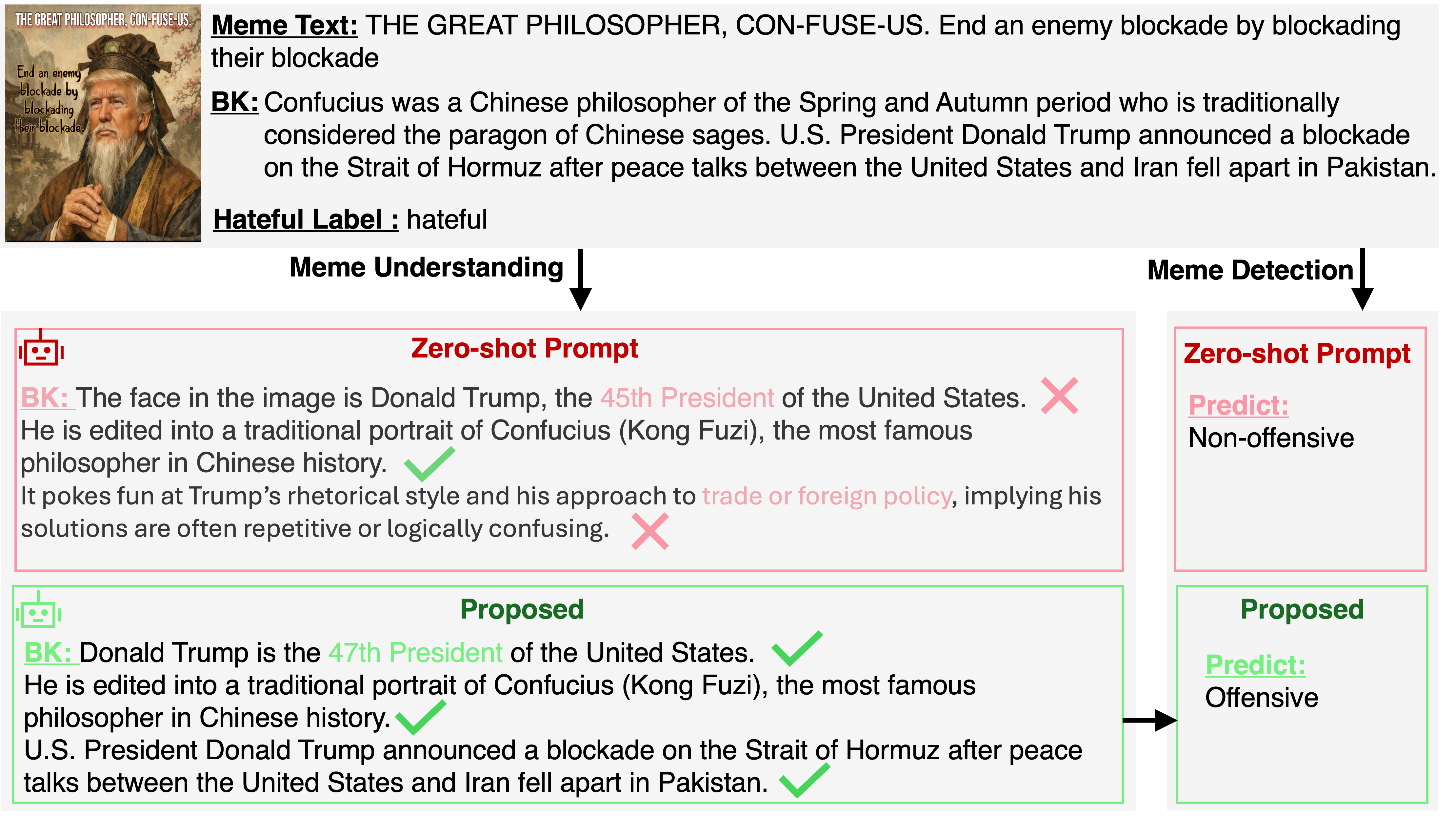}
    \caption{
    An example\protect\footnotemark{} illustrating the dynamic world knowledge gap in meme understanding. 
%The meme requires background knowledge beyond its visible image and text, including Confucius-related wordplay, and the Strait of Hormuz context. 
Direct zero-shot prompting misses the relevant context, whereas our framework recovers evidence-grounded background knowledge that supports correct downstream detection.
    }
    % \caption[Overview of automated meme understanding and detection.]{
    % Given an input meme\protect\footnotemark{} and its embedded text, our goal is to support automated meme understanding and detection. Compared with zero-shot prompting (\textcolor{pink}{in pink}), our pipeline recovers reference-aligned background knowledge and uses it as contextual evidence to support more accurate downstream prediction (\textcolor{green!60!black}{in green}).
    % }
    \label{fig:intro}
\end{figure}
\footnotetext{Source of the meme used in this figure: \url{https://knowyourmeme.com/}.}
%Yet, these contextual elements are often absent from the meme image and text themselves. Even with the rapid advancement of LVLMs, the relevant background knowledge may still be incomplete, outdated, or unavailable in a model's parametric memory depending on its knowledge cutoff time.\footnote{\url{https://en.wikipedia.org/wiki/Knowledge_cutoff}} This makes background knowledge acquisition a challenging yet important problem for meme understanding.
For example, the meme in Figure~\ref{fig:intro} requires {more than }recognizing the two depicted figures Donald Trump and Confucius. 
%while also interpreting the embedded textual wordplay. Specifically, 
The phrase ``CON-FUSE-US'' evokes both ``Confucius''  and the English word``confuse us'', 
creating a layered pun grounded in linguistic and cultural knowledge.
%, creating a layered pun that relies on both linguistic and cultural background knowledge.
%\footnote{Fun Fact or Additional background knowledge: The Mandarin Chinese pronunciation of Kǒngfūzǐ \zh{(孔夫子)} can be perceived as phonetically similar to ``confuse'' in English.}
However, the point of the pun can only be understood in relation to time-evolving real-world events, including Iran’s blockage of the Strait of Hormuz in February 2026 and the subsequent U.S. naval blockade after the failed Islamabad Talks.
%\footnote{\url{https://en.wikipedia.org/wiki/2026_Strait_of_Hormuz_crisis}; \url{https://en.wikipedia.org/wiki/2026_United_States_naval_blockade_of_Iran}}

Despite the inherently dynamic nature of memes, 
existing multimodal meme understanding methods remain limited in their use of background knowledge. 
Most methods built on vision-language models (VLMs) focus mainly on vision-language interactions (e.g., aligning regions with embedded text or capturing cross-modal incongruity)~\cite{DBLP:conf/aaai/SharmaASNA023, nguyen2025memeqa}, treating memes as closed multimodal reasoning problems. 
Moreover, they heavily rely on human annotated supervised data and the implicit parametric knowledge of VLMs~\cite{DBLP:conf/emnlp/HwangS23,park2025memeinterpret,park2024memeintent,nguyen2025memeqa}, making them less adaptable when required knowledge is missing, outdated, or newly emerging. Although one prior work~\cite{tripathi2026they} incorporates external knowledge bases (e.g., ConceptNet\footnote{https://conceptnet.io/} and Hatebase\footnote{https://hatebase.org/}), it relies on static  symbolic knowledge, limiting both updatability and human interpretability. 
%These limitations motivate methods that explicitly acquire and reason over diverse, human-interpretable background knowledge.

%As a result, the knowledge required to interpret a meme may be absent from the meme image and text themselves, outdated, or unreliable in the parametric memory of a pretrained vision-language model (VLM). This creates an open-world knowledge gap for automated meme understanding: the model must not only perceive the meme, but also acquire the missing background knowledge needed for interpretation.

{Meme detection builds on meme understanding, since reliable detection requires first interpreting the meaning and intention of the meme\citep{cao2022prompting, cao2023pro,lin2024goat,lin2025ask,liu2025mind,rizwan2026see,liu2026yes}.
However, existing detection methods share similar limitations: they mainly model multimodal interactions using supervised data~\cite{lee2021disentangling,agarwal2024mememqa}, and rely on parametric knowledge of VLMs~\cite{DBLP:journals/corr/abs-2506-08477,DBLP:journals/corr/abs-2510-08630} or large language models (LLMs)~\cite{DBLP:conf/www/LinLGMWY24, cao2022prompting,lin2025ask}.
These limitations make them less robust to emerging memes that require external commonsense, cultural, or time-sensitive background knowledge.
}
% Prior work has explored fusion-based multimodal architectures, prompting and in-context learning with large VLMs, and knowledge-enhanced methods for meme understanding ~\cite{lee2021disentangling,pramanick2021momenta,cao2022prompting,cao2023pro,lin2024goat,hee2025demystifying,chen2025adammeme,lin2025explainhm++}. 
%\crtext{Todo}
%Recent work has begun to examine the role of background knowledge in meme understanding. For example,  MemeIntent~\cite{park2024memeintent}, MemeQA~\cite{nguyen2025memeqa}, and MemeInterpret~\cite{park2025memeinterpret}, introduce human-annotated intents, rationales, or background knowledge for evaluating meme understanding. Other methods, such as CROSS-ALIGN+~\cite{tripathi2026they}, incorporate existing cultural resources, while MemeAgent~\cite{lin2025ask} uses agent-style question-answering discussions for social abuse detection. Despite these advances, these approaches often rely on task-specific training data, human-annotated background knowledge, predefined cultural resources, or the model's implicit parametric knowledge. 

\textbf{Our Approach}. 
This motivates our zero-shot framework, \textit{Query-Retrieve-Conclude}, which does not require task-specific supervised data.
Rather than relying solely on VLMs for meme understanding or detection, it retrieves open-web evidence to acquire the essential and up-to-date world knowledge needed to address the dynamic nature of memes. The framework consists of three stages. The \textit{Query} stage identifies missing knowledge by using reverse image search, caption generation, and question generation to produce search-oriented questions. 
The \textit{Retrieval} stage retrieves external evidence from the open-web for each question and generates evidence-grounded answers. 
The \textit{Conclude} stage synthesizes the resulting question-answer pairs into explicit background knowledge statements, following evidence-statement formulations in recent retrieval evaluation work~\cite{cao2025averimatec,akhtar2026ev2r}, and uses these statements as contextual evidence for downstream task. 
%This differs from direct zero-shot background knowledge generation, where a model is asked to infer the missing context in one step. 
%Such direct generation can produce plausible but unsupported knowledge, especially when the meme refers to recent or evolving events. 
%Our framework instead makes the missing-knowledge discovery process explicit: it first generates search-oriented questions, retrieves evidence for those questions, and only then synthesizes background knowledge statements for detection.

%Crucially, rather than treating meme interpretation as an isolated task, we inject these synthesized statements as dynamic contextual evidence into downstream detection. 
%By separating open-world knowledge acquisition from downstream detection, our pipeline explicitly stabilizes predictions on complex reasoning tasks, such as hatefulness, misogyny, and sarcasm detection, without requiring task-specific fine-tuning. To the best of our knowledge, we are the first to frame meme interpretation under dynamic, evolving contexts as a query-driven, retrieval-grounded knowledge discovery problem.
Beyond meme understanding, we further apply our framework to meme detection to further validate the utility and quality of generated background knowledge. The same pipeline produces evidence-grounded knowledge statements that help reveal the meme’s intended meaning, which are then used as contextual evidence for detection. This allows detection decisions to be grounded not only in visual-textual interactions, but also in the external world knowledge needed to interpret dynamic events, cultural references, and evolving online discourse.

To validate our proposed method, we benchmark it on both meme understanding and meme detection datasets. For meme understanding, in addition to existing benchmarks, we introduce \textit{KYM}, a highly curated diagnostic benchmark of recent memes from 2024 to 2026 that capture volatile internet phenomena. KYM further provides external background knowledge annotations, enabling evaluation under strict temporal distribution shifts and testing whether models can interpret memes grounded in rapidly evolving real-world events and online discourse.
Experiments on \textbf{three meme understanding datasets} and \textbf{five meme detection datasets} demonstrate the effectiveness of our proposed method.

In summary, our contributions are as follows:
\begin{itemize}
   \item We identify the dynamic world-knowledge gap in existing supervised and zero-shot meme understanding methods.
   \item We propose \textit{Query-Retrieve-Conclude}, a zero-shot framework that retrieves open-web evidence and synthesizes grounded background knowledge for meme understanding and detection.
   \item We conduct extensive experiments on three meme understanding datasets, including \textit{KYM}, a curated benchmark for recent memes tied to evolving real-world events, and five meme detection datasets.
\end{itemize}

% A meme is not simply an image, a caption, or a set of external facts in isolation; rather, its meaning emerges from the interaction among visual cues, textual cues, and culturally situated background knowledge~\cite{cao2022prompting,cao2023pro,park2024memeintent,nguyen2025memeqa}.
% As \cite{nguyen2025memeqa} observed, "Automated meme understanding requires systems to demonstrate fine-grained visual recognition, commonsense reasoning, and extensive
% cultural knowledge." \footnote{We position our proposed automated meme understanding and detection framework as complementary to the dimensions of holistic meme understanding discussed in~\cite{nguyen2025memeqa}.} 

\section{Related Work}
%meme understanding and detection tasks
\begin{figure*}[t!]
    \centering
    \includegraphics[width=0.98\textwidth, trim={0cm 0cm 0cm 0cm}, clip]{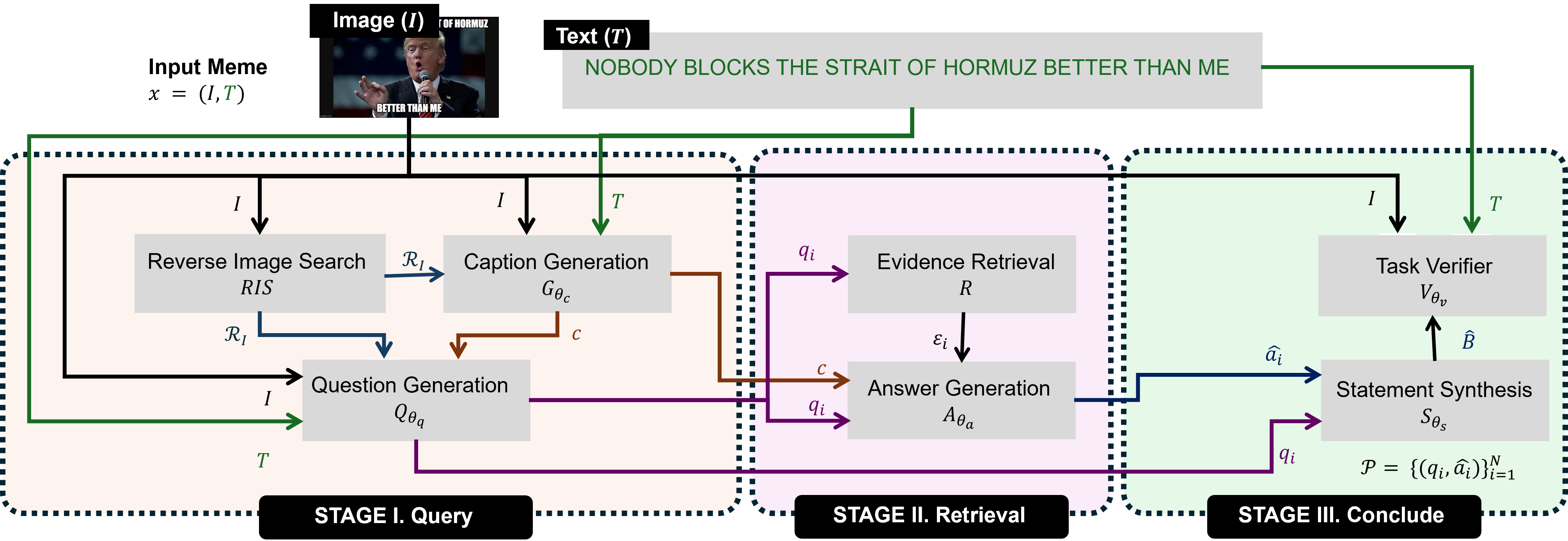}
    \caption{Overview of the proposed query-driven, retrieval-grounded framework for automated meme understanding and detection. 
Given a meme image $I$ and text $T$, our proposed framework is organized into three stages. 
Stage I (\textit{Query}) identifies missing knowledge through reverse image search, caption generation, and question generation. 
Stage II (\textit{Retrieval}) retrieves external evidence for each generated question and produces evidence-grounded answers. 
Stage III (\textit{Conclude}) synthesizes the question--answer pairs into explicit background knowledge statements $\hat{B}$ and uses them as contextual evidence for downstream meme detection.
}
    \label{fig:pipeline}
\end{figure*}

%\paragraph{Multimodal Meme Understanding}
%Multimodal memes is a prevelant communication message 

%\paragraph{Knowledge Integration} 
Knowledge integration has proven to be effective in NLP tasks such as factual reasoning~\cite{petroni2019language,cao2025averimatec,akhtar2026ev2r}. For multimodal meme understanding, recent studies have shown that external knowledge can enhance meme understanding and detection. For example, PromptHate~\cite{cao2022prompting} and Pro-Cap~\cite{cao2023pro} utilize pretrained vision-language models to obtain contextual and cultural information for hateful meme analysis. More recent automated meme understanding studies, such as MemeIntent~\cite{park2024memeintent} and MemeQA~\cite{nguyen2025memeqa}, collect background knowledge for existing datasets and fine-tune large multimodal models to support more holistic meme understanding. Additionally, \cite{tripathi2026they} provides cultural grounding by integrating knowledge from existing meme datasets.
Although these studies have enhanced understanding and detection, respectively, they remain limited when applied to recent or emerging memes whose meanings depend on timely events and evolving internet culture.
More importantly, they do not explicitly model \emph{what knowledge is missing} before attempting to explain or classify a meme. 
As a result, background knowledge is often treated as either an annotation, a retrieved resource, or a direct model output, rather than as the result of an explicit knowledge-gap identification and evidence acquisition process.

% AVERIMATEC~\cite{cao2025averimatec} and \cite{han2026beyond} recently demonstrated that external knowledge can enhance fact-checking and misinformation detection.

%  \paragraph{A Brave New World: Automated Meme Understanding and Detection}

% A meme is not simply an image, a caption, or a set of external facts in isolation; rather, its meaning emerges from the interaction among visual cues, textual cues, and culturally situated background knowledge~\cite{cao2022prompting,cao2023pro,park2024memeintent,nguyen2025memeqa}.
% As \cite{nguyen2025memeqa} observed, "Automated meme understanding requires systems to demonstrate fine-grained visual recognition, commonsense reasoning, and extensive
% cultural knowledge." \footnote{We position our proposed automated meme understanding and detection framework as complementary to the dimensions of holistic meme understanding discussed in~\cite{nguyen2025memeqa}.} 

% Although \cite{tripathi2026they} has attempted to a systematic integration of cultural knowledge for provides cultural grounding by integrating knowledge from existing datasets. Despite strong progress, their approach rely on annotated data static knowledge making it difficult to handle newly emerging memes whose meanings depend on evolving cultural or event-specific context. 

\section{Methodology}
\label{sec:proposed_pipeline}

We propose a query-driven, retrieval-grounded, and conclusion-oriented framework for meme understanding. 
This design is motivated by how humans interpret unfamiliar memes. 
When encountering a meme whose meaning is not immediately clear, a reader may first identify unfamiliar entities, phrases, or visual references; then search for relevant information; and finally combine the retrieved knowledge with the meme content to infer its intended meaning or social implication. 
As shown in Figure~\ref{fig:pipeline}, the framework consists of three structured stages: 
\textit{Query}, \textit{Retrieve}, and \textit{Conclude}. 
Rather than directly prompting a large VLM to explain or classify a meme, 
the novelty of our framework lies in treating background knowledge acquisition as a structured reasoning process: query--retrieve--conclude process. 
Rather than as a single-step generation problem or a direct classification prompt, we first identifying knowledge gaps, then retrieving external evidence, and finally synthesizing the retrieved information into task-relevant background knowledge for detection.

\subsection{Stage I: Query}
The first stage identifies the missing knowledge required to interpret the meme. 
To reduce the risk of prematurely fabricating background knowledge from parametric memory, this stage asks the model to formulate what needs to be known before the meme can be understood, thereby encouraging evidence-seeking behavior.
Given the meme image $I$ and text $T$, the goal is to generate a compact set of search-oriented questions 
$Q=\{q_1,\ldots,q_N\}$ that guide evidence retrieval. 
Unlike direct explanation-based prompting, this stage does not ask the model to immediately infer the meme's meaning. 
We first apply reverse image search (RIS) to the meme image to obtain visually associated web context, denoted as $\mathcal{R}_I = \mathrm{RIS}(I)$. 
Next, we generate a visually grounded caption $c$ from the image, text, and reverse-image-search context. 
The caption generator uses $\mathcal{R}_I$ only as auxiliary visual grounding, helping resolve ambiguous entities, symbols, or scenes while preserving a literal description of the meme. 
Finally, the question generator produces a set of search-oriented questions, 
\begin{equation}
    Q = Q_{\theta_q}(I,T,c,\mathcal{R}_I).
\end{equation}
Each question targets a potential knowledge gap, such as the identity of a depicted person, the meaning of a phrase, the background of a public event, or the relationship between the image and the text. 
The output of this stage is therefore not a final interpretation, but a structured query plan for external knowledge acquisition.

\subsection{Stage II: Retrieve}
This retrieve stage is central to the open-world nature of our framework. 
Because memes often refer to recent events, evolving cultural references, or platform-specific discourse, relying only on pretrained parametric knowledge can be insufficient \cite{valensise2021entropy}. 
Recognizing this gap, the second stage is designed to retrieve external evidence at inference time, such that the framework can adapt to emerging memes whose meanings may not be fully encoded in the model parameters.
For each question $q_i \in Q$, the evidence retrieval module searches the web and returns a set of relevant passages, i.e., $\epsilon_i = R(q_i),$
where $R(\cdot)$ denotes the web search and evidence retrieval module. 
The full retrieved evidence set is $\epsilon=\{\epsilon_i\}_{i=1}^{N}$.
Given each question $q_i$, its retrieved evidence $\epsilon_i$, and the generated caption $c$, the answer generator produces a concise evidence-grounded answer:
\begin{equation}
    \hat{a}_i = A_{\theta_a}(c,q_i,\epsilon_i).
\end{equation}
The caption provides visual context, while the retrieved passages provide external textual grounding. 
The answer generator is constrained to use only the retrieved evidence, ensuring that the resulting answer is grounded in externally accessible information rather than unsupported model assumptions.

\subsection{Stage III: Conclude}
The conclusion stage serves two purposes: (i) to produce interpretable background knowledge statements that can be directly evaluated for meme understanding; and (ii) to use these statements as task-level evidence for downstream detection. 
This dual role allows the framework to connect meme understanding and meme detection within a single pipeline.
The output from the previous two stages create a set of question-answer pairs, $\mathcal{P} = \{(q_i,\hat{a}_i)\}_{i=1}^{N}$, which are first transformed into declarative background knowledge statements:
\begin{equation}
    \hat{B} = S_{\theta_s}(\mathcal{P}),
\end{equation}
where $S_{\theta_s}$ denotes the statement synthesis module and 
$\hat{B}=\{\hat{b}_1,\ldots,\hat{b}_M\}$ denotes the synthesized background knowledge. 
This step converts question-specific answers into compact statements that can be evaluated against human-annotated background knowledge and reused as contextual evidence for any task.
For example, given the meme image $I$, text $T$, and synthesized background knowledge $\hat{B}$, we can design the verifier predicts the binary label for each task $s \in \mathcal{S}$:
\begin{equation}
    \hat{y}^{(s)} = V^{(s)}_{\theta_v}(I,T,\hat{B}), 
    \quad \hat{y}^{(s)} \in \{0,1\}.
\end{equation}
Unlike direct zero-shot detection, the verifier is provided with explicit background knowledge that explains relevant entities, events, phrases, or image-text associations. 
This enables the prediction to be grounded in retrieved evidence rather than relying only on the model's implicit knowledge.

\section{Experiment}
\label{experiment}

\begin{figure}[t!]
    \centering
    \includegraphics[
        width=\linewidth,
        trim={0cm 0cm 0cm 0cm},
        clip
    ]{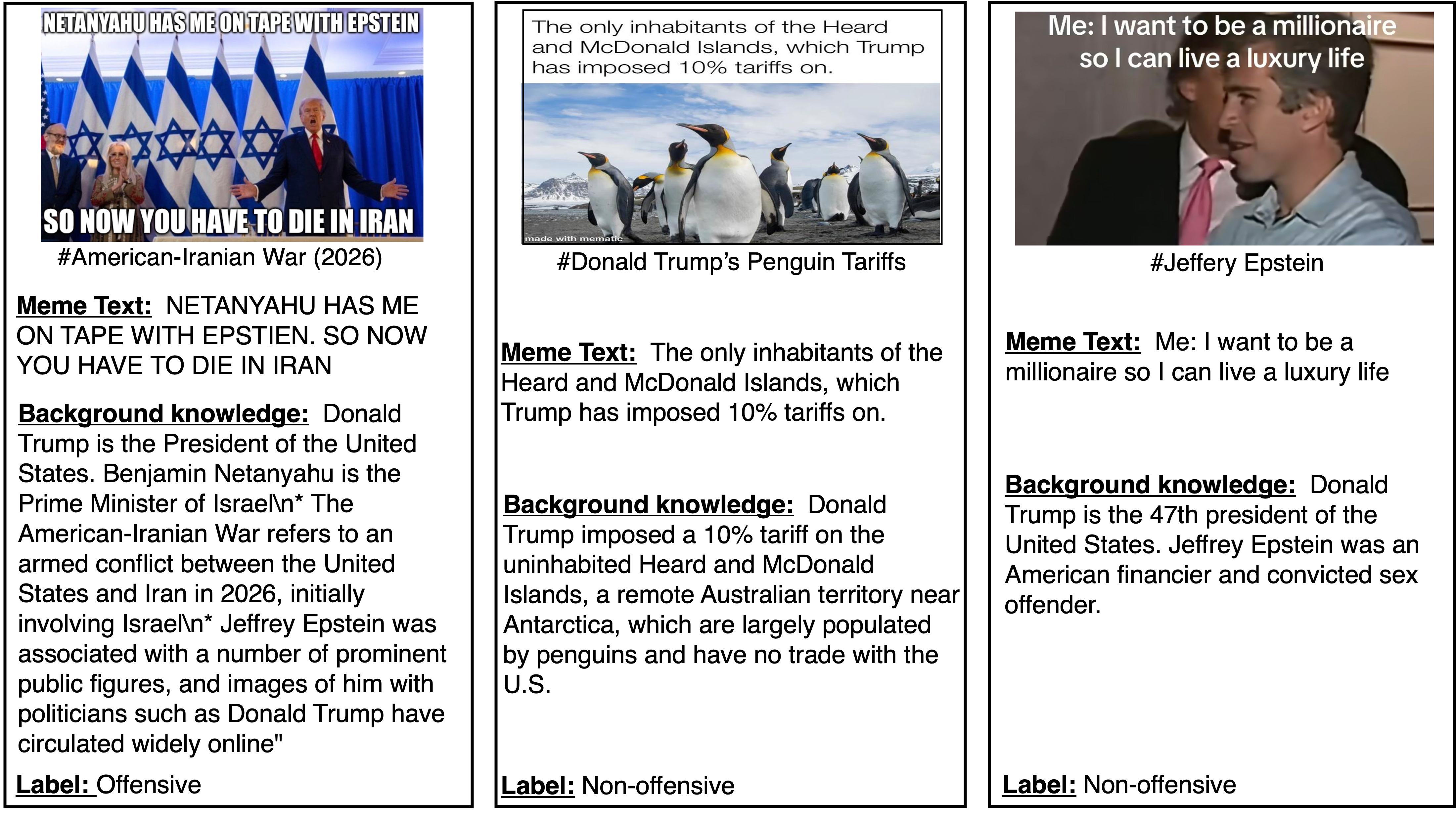}
    \caption{Example memes from the curated KYM dataset, collected from Know Your Meme, series on \#2026 American-Iranian War; \#Donald Trump's Penguin Tariffs; \#Jeffery Epstein (from left to right).}
    \label{fig:kym}
\end{figure}

\subsection{Implementation Details}
\label{datasets}
%understanding and detection datasets
%Ev2R evaluation and human alignment check
\textbf{Datasets.}
For meme understanding, we use the existing MemeIntent~\cite{park2024memeintent} and MemeInterpret~\cite{park2025memeinterpret} datasets, and further curate a new KYM dataset to evaluate whether our framework can handle newly emerging memes. As shown by the examples in Figure~\ref{fig:kym}, each meme in the KYM dataset is annotated with background knowledge, intent and an offensiveness label. Overall we annotated 100 memes scraped from \emph{Know Your Meme}\footnote{\url{https://knowyourmeme.com/}}, covering meme series from 2024 to 2026. Specifically, these memes are centered on recent and culturally situated topics, such as the \emph{American--Iranian War}, \emph{Jeffrey Epstein}, \emph{Donald Trump's interest in purchasing Greenland}, and \emph{the United States of America}.  Which often depend on evolving political events, public discourse, and cultural references, thus they provide a challenging testbed for evaluating whether our framework can recover up-to-date background knowledge for meme interpretation. 

% \footnote{Details of its curation can be found in Appendix~\ref{app:kym}.}

For downstream meme detection task, inspired by GOAT-Bench~\cite{lin2024goat}, we evaluate our framework across five tasks related to social abuse and nuanced meme interpretation: hatefulness, misogyny, offensiveness, sarcasm, harmfulness, and humor. Table~\ref{tab:detect_stats} summarizes the label distribution and dataset size for each task.

\begin{table}[t]
\centering
\scriptsize
\setlength{\tabcolsep}{2pt}
\renewcommand{\arraystretch}{1}
%\resizebox{\columnwidth}{!}{%
\begin{tabular}{llccc}
\toprule
\textbf{Task} & \textbf{Source} & \multicolumn{2}{c}{\textbf{Label Distribution}} & \textbf{Total} \\
\cmidrule(lr){3-4}
 & & \textbf{Label} & \textbf{Count} & \\
\midrule
\multirow{2}{*}{Hatefulness}
& \multirow{2}{*}{MemeInterpret}
& Hateful & 490 & \multirow{2}{*}{1000} \\
& & Non-hateful & 510 & \\
\midrule

\multirow{2}{*}{Misogyny}
& \multirow{2}{*}{MAMI}
& Misogynistic & 500 & \multirow{2}{*}{1000} \\
& & Non-misogynistic & 500 & \\
\midrule

\multirow{2}{*}{Offensiveness}
& \multirow{2}{*}{MultiOFF}
& Offensive & 303 & \multirow{2}{*}{743} \\
& & Non-offensive & 440 & \\
\midrule

\multirow{2}{*}{Sarcasm}
& \multirow{2}{*}{MSD}
& Sarcastic & 910 & \multirow{2}{*}{1820} \\
& & Non-sarcastic & 910 & \\
\midrule

\multirow{2}{*}{Harmfulness}
& \multirow{2}{*}{\makecell{Harm-C,\\Harm-P}}
& Harmful & 444 & \multirow{2}{*}{1063} \\
& & Non-harmful & 619 & \\
%\midrule

% \multirow{2}{*}{Humor}
% & \multirow{2}{*}{PrideMM}
% & Humour & 478 & \multirow{2}{*}{735} \\
% & & Non-humour & 257 & \\
\bottomrule
\end{tabular}%
%}
\caption{Dataset statistics for meme detection tasks.}
\label{tab:detect_stats}
\end{table}

\begin{table}[t]
\centering
\small
\setlength{\tabcolsep}{2pt}
\begin{tabular}{ll|ccc}
\toprule
\multicolumn{2}{c|}{\textbf{Model}}
& \multicolumn{3}{c}{\textbf{Datasets}} \\
\cmidrule(lr){1-2}
\cmidrule(lr){3-5}

\makecell[c]{\textbf{QA-pairs}} 
& \makecell[c]{\textbf{BKS}}
& \textbf{KYM}
& \textbf{MemeIntent}
& \textbf{MemeInterpret} \\ 
\midrule

\multicolumn{5}{c}{\textbf{Zero-shot Generated BKS}} \\
\midrule
-- & Qwen3
& 0.46 & 0.66 & 0.73 \\

-- & Gemma3
& 0.59 & 0.65 & 0.66 \\

\midrule
\multicolumn{5}{c}{\textbf{**Proposed}} \\
\midrule
Gemma3 & Gemma3
&0.65  &0.70  & 0.74 \\

Qwen3 & Qwen3
&0.78 &0.76 & 0.79 \\

Gemma3 & Qwen3
&0.68  &0.73  & 0.76 \\

Qwen3 & Gemma3
& 0.78 &0.75 & 0.78 \\

\bottomrule
\end{tabular}
%}
\caption{Recall scores of zero-shot generated background knowledge and QA-pair-converted knowledge statements against ground-truth background knowledge. Qwen3-VL-32B (Qwen3) and Gemma-3-12B-IT (Gemma3) are used interchangeably for QA-pair generation and knowledge-statement generation.}
\label{tab:understand_results}
\end{table}

\begin{table*}[t]
\centering
\scriptsize
\setlength{\tabcolsep}{7pt}
% \resizebox{\textwidth}{!}{%
\begin{tabular}{
>{\centering\arraybackslash}p{0.20cm}
l|*{12}{c}
}
\toprule
\multicolumn{2}{c|}{\textbf{Model}}
& \multicolumn{12}{c}{\textbf{Detection Task}} \\
\cmidrule(lr){1-2}
\cmidrule(lr){3-14}

& \textbf{Detection}
& \multicolumn{2}{c}{\makebox[1.24cm][c]{\textbf{Hatefulness}}}
& \multicolumn{2}{c}{\makebox[1.24cm][c]{\textbf{Misogyny}}}
& \multicolumn{2}{c}{\makebox[1.24cm][c]{\textbf{Offensiveness}}}
& \multicolumn{2}{c}{\makebox[1.24cm][c]{\textbf{Sarcasm}}}
& \multicolumn{2}{c}{\makebox[1.24cm][c]{\textbf{Harmfulness}}}
& \multicolumn{2}{c}{\makebox[1.24cm][c]{\textbf{Overall}}} \\
\cmidrule(lr){3-4}
\cmidrule(lr){5-6}
\cmidrule(lr){7-8}
\cmidrule(lr){9-10}
\cmidrule(lr){11-12}
\cmidrule(lr){13-14}

&
& \textbf{Acc.} & \textbf{F1}
& \textbf{Acc.} & \textbf{F1}
& \textbf{Acc.} & \textbf{F1}
& \textbf{Acc.} & \textbf{F1}
& \textbf{Acc.} & \textbf{F1}
& \textbf{Acc.} & \textbf{F1} \\
\midrule

\multicolumn{14}{c}{\textbf{Zero-shot without BKS}}\\
\midrule

\multirow{3}{*}{\rotatebox[origin=c]{90}{\tiny Vanilla}}
& LLaVA-1.5-7B
& 0.54 & 0.45
& 0.59 & 0.52
& 0.58 & 0.57
& 0.50 & 0.34 
& 0.61 & 0.54 
& 0.56 & 0.48  \\

& Qwen3-VL-8B
& 0.63 & 0.61
& 0.64 & 0.62
& 0.60 & 0.59
& 0.61 & 0.60 
& 0.63 & \textbf{0.61} 
& 0.62 & 0.61 \\

& Gemma3-12B
& 0.67 & 0.67 
& 0.73 & 0.72
& 0.62 & 0.61
& 0.63 & 0.62 
& 0.65 & 0.64 
& 0.67 & 0.65 \\ 

\midrule

\multirow{3}{*}{\rotatebox[origin=c]{90}{\tiny MemeAgent}}
& LLaVA-1.5-7B
&0.55  &0.43 
& 0.60 & 0.45
&\uline{0.60}  & 0.45
& 0.56 & 0.42
&0.58  &0.46 
&0.58  & 0.44 \\

& \rule{0pt}{1.2em}Qwen3-VL-8B
&0.64  &0.60 
& \uline{0.67} & \textbf{0.65}
&\uline{0.62}  & 0.41
& 0.64 & 0.60
&0.62  &0.59 
&\uline{0.64}  &0.57 \\

& \rule{0pt}{1.2em}Gemma3-12B
&0.66  &0.63 
& 0.61 & 0.49 
&0.60  & 0.56
&0.61  &0.65  
&0.62  &0.58 
&0.62  &0.59  \\ 
\midrule

\multirow{3}{*}{\rotatebox[origin=c]{90}{\tiny MiND}}
& LLaVA-1.5-7B
&0.54  & \textbf{0.60} 
&0.57  &0.50  
&0.59  & 0.56 
&0.55  & 0.47 
&0.60  & 0.53 
&0.58  & 0.44   \\

& Qwen3-VL-8B
&0.63  &0.60
&\textbf{0.68}  &\uline{0.64}    
& 0.61 &0.58  
&0.63  &0.59 
&0.64  & 0.60  
&\uline{0.64}  & 0.57   \\

& Gemma3-12B
& 0.66 & 0.65 
& 0.70 & 0.68
& 0.61 & 0.59
& 0.62 & 0.61
& 0.64 & 0.62
& 0.65 & 0.63 \\

\midrule

\multicolumn{14}{c}{\textbf{With Zero-shot Generated BKS}}\\
\midrule

& LLaVA-1.5-7B
& 0.56 & 0.42
& \uline{0.61} & \uline{0.55}
& \textbf{0.62} & \uline{0.60}
& \uline{0.61} & \uline{0.56} 
& \uline{0.63} & \uline{0.55} 
& \uline{0.61} & \uline{0.54}  \\

& Qwen3-VL-8B
& \uline{0.65} & 0.60 
&0.65 & 0.61
& \uline{0.62} & \uline{0.61}
& \uline{0.63} & \textbf{0.63} 
&\uline{0.65} & \uline{0.60}
&\uline{ 0.64} & \uline{0.61}  \\

& Gemma3-12B
& 0.69 & 0.66
& 0.75 & 0.74 
& 0.66 & 0.64
& 0.65 & 0.64 
& 0.67 & 0.66
& 0.68 & 0.67  \\

\midrule
\multicolumn{14}{c}{**\textbf{Proposed}} \\
\midrule

& LLaVA-1.5-7B
& \textbf{0.61} & \uline{0.59}
& \textbf{0.63} & \textbf{0.60}
& \textbf{0.62} & \textbf{0.61}
& \textbf{0.62} &\textbf{0.59}
& \textbf{0.64} & \textbf{0.60} 
& \textbf{0.62} & \textbf{0.60}  \\

& Qwen3-VL-8B
& \textbf{0.66} & \textbf{0.62}
& \uline{0.67} & \uline{0.64}
& \textbf{0.64} & \textbf{0.62}
& \textbf{0.65} & \uline{0.62}
& \textbf{0.66} & \uline{0.61}
& \textbf{0.66} & \textbf{0.62}  \\

& Gemma3-12B
& \textbf{0.71} & \textbf{0.70}
& \textbf{0.80} & \textbf{0.79}
& \textbf{0.67} & \textbf{0.67}
& \textbf{0.69} & \textbf{0.68}
& \textbf{0.70} & \textbf{0.70} 
& \textbf{0.71} & \textbf{0.71}  \\

\bottomrule
\end{tabular}
% }
\caption{Experimental results (Accuracy and F1) across five meme detection tasks. Results evaluate zero-shot detection without background knowledge, dedicated baselines (MemeAgent, MiND), zero-shot generated BKS, and our proposed \textit{Query-Retrieve-Conclude}. Best results for each model backbone are in \textbf{bold}; second-best are \uline{underlined}.}
% \caption{Experimental results on meme detection tasks. We compare zero-shot detection without background knowledge, MemeAgent, MIND, detection with zero-shot generated background knowledge, and detection with background knowledge from our proposed framework.}
\label{tab:detect_results}
\end{table*}

\textbf{Evaluation.}
We evaluate both evidence recovery quality and downstream meme detection performance. Recent studies~\cite{cao2025averimatec,akhtar2026ev2r} show that reference-based evidence evaluation with large language models (LLMs) aligns best with human judgments. We extend this idea to our multimodal meme understanding setting, where retrieved web evidence may include both text and images. Recall that each generated QA pair is converted into an evidence statement. We then perform reference-based evaluation by comparing these generated evidence statements with the ground-truth evidence annotations. We exploit Gemini-3.1-Flash \footnote{\url{https://deepmind.google/models/model-cards/gemini-3-1-flash-image/}} as the scoring model for the this evaluation inspired by its power \cite{cao2025averimatec,akhtar2026ev2r}. We then report
evidence recall, defined as the percentage of ground-truth evidence instances successfully retrieved. The prompts used for QA-pair-to-statement conversion and evidence evaluation are provided in Appendix~\ref{app:prompt-QA_convert} and Appendix~\ref{app:prompt-evidence_eval}, respectively. Additionally, to assess the reliability of the reference-based evaluation, we compare the resulting scores with human judgments obtained from independent raters (details in Appendix ~\ref{app:align-evidence}). % to do: report the scores with algnment scores 

For downstream meme detection, we evaluate how well the recovered evidence supports prediction across the target tasks, using Accuracy and Macro-averaged F1 Score as the main evaluation metrics. We additionally conduct a human-alignment check, referred to as \emph{Decision Support}, to examine whether the predicted background knowledge contains sufficient information to justify the annotated detection labels (details in Appendix~\ref{app:align-detect}). 
% to do: report the scores

\textbf{Models.}
We use different models for meme understanding and meme detection. For meme understanding, \textbf{Qwen3-VL-32B-Instruct}\cite{bai2025qwen3} and \textbf{Gemma-3-27B-it}\cite{gemma_team_gemma_2025} are used. For reference-based evaluation, we use Gemini-3.1-Flash. For meme detection, we evaluate three open-source LVLMs: \textbf{llava-1.5-7b-hf}\cite{liu2023visual}, \textbf{Qwen3-VL-8B-Instruct}\cite{bai2025qwen3}, and \textbf{Gemma-3-12B-it}\cite{kamath2025gemma}.

\subsection{Baselines}
To contextualize the performance of our proposed framework, we compare against zero-shot baselines that with and without background knowledge, which are:
\begin{itemize}
    \item \textbf{Zero-shot inference without background knowledge.} 
    We evaluate meme detection using only the image and embedded text, without any generated background knowledge. This setting includes vanilla zero-shot inference, MemeAgent~\cite{lin2025ask}, an agent-based zero-shot baseline that conducts a discussion over the meme before producing the final prediction, and MiND~\cite{liu2025mind}, a retrieval-augmented zero-shot baseline that does not update model parameters but uses similar memes from a retrieval pool to derive task-level insights for final inference.
    \item \textbf{Detection with zero-shot generated background knowledge.} Following MemeInterpret~\cite{park2025memeinterpret}, we prompt Qwen3-VL-32B to generate background knowledge directly from the meme image and text, and then use the generated knowledge as additional context for downstream detection~\footnote{The prompts used for zero-shot knowledge generation and detection are shown in Appendix~\ref{app:prompt-zsbks} and Appendix~\ref{app:prompt-detect} respectively}.
\end{itemize}

\subsection{Results and Discussion}
The evaluation results for automated meme understanding and downstream detection are presented in Table~\ref{tab:understand_results} and Table~\ref{tab:detect_results}, respectively.  Our analysis focuses on how open-world knowledge acquisition fills critical reasoning gaps where closed parametric baselines fail.

\paragraph{Meme Knowledge Recovery Analysis.}
As reported in Table~\ref{tab:understand_results}, zero-shot models that rely entirely on internal parametric parameters struggle to reconstruct accurate background knowledge statements (BKS), particularly on newly emerging contexts. Under the zero-shot baseline, Qwen3 achieves an evidence recall score of only 0.46 on the curated KYM dataset, demonstrating that even large-scale VLMs fail to encapsulate highly dynamic, temporally shifting sociopolitical events (2024–2026).

In contrast, our proposed multi-stage framework consistently improves evidence recall across all evaluated benchmarks. When pairing Qwen3 as both the QA-pair generator and the BKS synthesizer, our framework improves recall on the KYM dataset from 0.46 to {0.78} (+32\% absolute increase). This performance gain is similarly reflected across established benchmarks, with recall advancing from 0.66 to 0.76 on MemeIntent, and from 0.73 to 0.79 on MemeInterpret.
Crucially, the cross-model evaluation (e.g., using Qwen3 for query generation and Gemma3 for statement synthesis, yielding 0.78 on KYM) reveals that the performance bottleneck in meme understanding lies not in the text synthesis stage, but in the precision of the initial search query. 
%By explicitly decomposing visual-textual incongruities into targeted open-world questions rather than predicting flat summaries, our framework bridges the temporal knowledge gap regardless of the downstream text model's parametric capacity.

\paragraph{Downstream Detection Performance.}
The empirical results across the five downstream detection tasks (Table~\ref{tab:detect_results}) confirm that the quality of recovered background knowledge translates directly into reasoning accuracy. Under the zero-shot without BKS baseline, vanilla models exhibit a severe performance ceiling; Gemma3-12B yields an overall F1 of 0.65, while smaller architectures like LLaVA-1.5-7B achieve just 0.48.

Iterative multi-agent reasoning (MemeAgent) and standard instance-level retrieval (MiND) offer marginal adjustments but do not break this ceiling. For instance, on the highly nuanced task of Sarcasm detection, MemeAgent decreases LLaVA-1.5's performance or plateaus Gemma3's performance (0.65 F1) because multi-agent debates over unencoded open-world context inevitably amplify parametric hallucinations. Similarly, while MiND utilizes visual similarity to anchor its predictions, it yields zero architectural mechanism to capture rapid temporal distribution shifts, keeping Gemma3's overall F1 constrained to 0.63.

Augmenting models with standard zero-Shot Generated BKS yields modest baseline improvements (e.g., raising Gemma3's overall F1 to 0.67), but it introduces severe structural noise. As observed in the LLaVA-1.5 configuration, zero-shot background knowledge actually degrades performance on Hatefulness (F1 drops from 0.45 to 0.42) because direct contextual generation induces severe semantic anchoring traps (further analyzed in Section~\ref{ss:case_study}).

Our proposed framework completely bypasses these degradation risks. Grounding downstream classifiers with our open-world knowledge framework pushes Gemma3-12B to a state-of-the-art accuracy of {0.71} and an F1 score of {0.71}. Most notably, our framework secures massive, absolute F1 leaps on culturally dense fields: +{0.07} in Misogyny (0.79 vs. 0.72) and +{0.06} in Sarcasm (0.68 vs. 0.62) compared to vanilla execution. 

% understanding table: recall scores comparison, our method most obvivous gains on kym 
% 

% \paragraph{Are questions really needed?}
% Yes and No. 
% % Yes for: compare ours and MemeAgent with no bks, and zero-shot generated bks. 
% % No for: MemeAgent only ask surface level questions, whereas ours mirrors human balabala..
\begin{table}[t!]
\centering
\small
\setlength{\tabcolsep}{3.5pt}
\renewcommand{\arraystretch}{1.1}
\resizebox{\columnwidth}{!}{%
\begin{tabular}{l | c c | c c | c c}
\toprule
\textbf{Framework Configuration} & \multicolumn{2}{c|}{\textbf{Hatefulness}} & \multicolumn{2}{c|}{\textbf{Sarcasm}} & \multicolumn{2}{c}{\textbf{Overall}} \\
 & \textbf{Acc.} & \textbf{F1} & \textbf{Acc.} & \textbf{F1} & \textbf{Acc.} & \textbf{F1} \\
\midrule
\textbf{Ours (Full Framework)} & \textbf{0.75} & \textbf{0.74} & \textbf{0.72} & \textbf{0.71} & \textbf{0.74} & \textbf{0.74} \\
\quad \textit{w/o Stage I: (Direct Retrieval)} & 0.69 & 0.66 & 0.67 & 0.65 & 0.69 & 0.67 \\
\quad \textit{w/o Stage II: (Parametric Only)} & 0.66 & 0.62 & 0.65 & 0.62 & 0.66 & 0.62 \\
\quad \textit{w/o Stage III: (Raw QA)} & 0.72 & 0.71 & 0.69 & 0.68 & 0.71 & 0.70 \\
\quad \textit{w/o Visual Grounding Context ($c$ \& $\mathcal{R}_I$)} & 0.65 & 0.61 & 0.63 & 0.60 & 0.64 & 0.61 \\
\bottomrule
\end{tabular}}
\caption{Ablation analysis across representative tasks using the Qwen3-VL-32B backbone. Configurations isolate down-stream performance shifts when structural blocks are systematically removed from our query-driven, open-world knowledge acquisition framework.}
\label{tab:ablation_results}
\end{table}
\paragraph{Ablation Study.} 
To isolate the exact architectural contributions of our \textit{Query-Retrieve-Conclude} framework, we systematically remove key components using the primary Qwen3-VL-32B backbone. Eliminating the structured Question Generation phase (\textit{w/o Stage I}) forces the system to pull open-world web evidence based purely on surface-level text or visual keywords. This configuration causes the overall F1 score to drop significantly from 0.74 to 0.67, confirming that direct query scraping fails to capture the implicit, multi-layered ironies embedded in evolving memes. Bypassing external web searches entirely and forcing a reliance on internal parametric parameters (\textit{w/o Stage II}) results in an even more severe performance collapse, dropping the Hatefulness F1 score to 0.62. This drop highlights that even a large-scale 32B foundation model cannot reliably infer shifting or highly volatile real-world contexts without live information retrieval. Finally, feeding uncurated, raw question-answer fragments directly to the downstream discriminator rather than using our statement synthesis module (\textit{w/o Stage III}) limits contextual efficiency and introduces structural token noise, which decreases the F1 score by a distinct 0.04 points. This systematic degradation across all ablation states demonstrates that each step in our framework is essential for accurately identifying knowledge gaps and protecting downstream models from semantic anchoring errors.

\begin{table*}[t]
\centering
\renewcommand{\arraystretch}{1.12}
\setlength{\tabcolsep}{3pt}
\scriptsize
%\resizebox{\textwidth}{!}{%
\begin{tabular}{|>{\centering\arraybackslash}p{0.45\textwidth}
                |>{\raggedright\arraybackslash}p{0.29\textwidth}
                |>{\raggedright\arraybackslash}p{0.16\textwidth}|}
\hline

\multicolumn{1}{|c|}{\textbf{Meme}}
& \multicolumn{1}{c|}{\textbf{Proposed}}
& \multicolumn{1}{c|}{\textbf{Zero-shot Baseline}}
\\
\hline

% =========================================================
% Meme 1
% =========================================================
\contextmemeimgtext{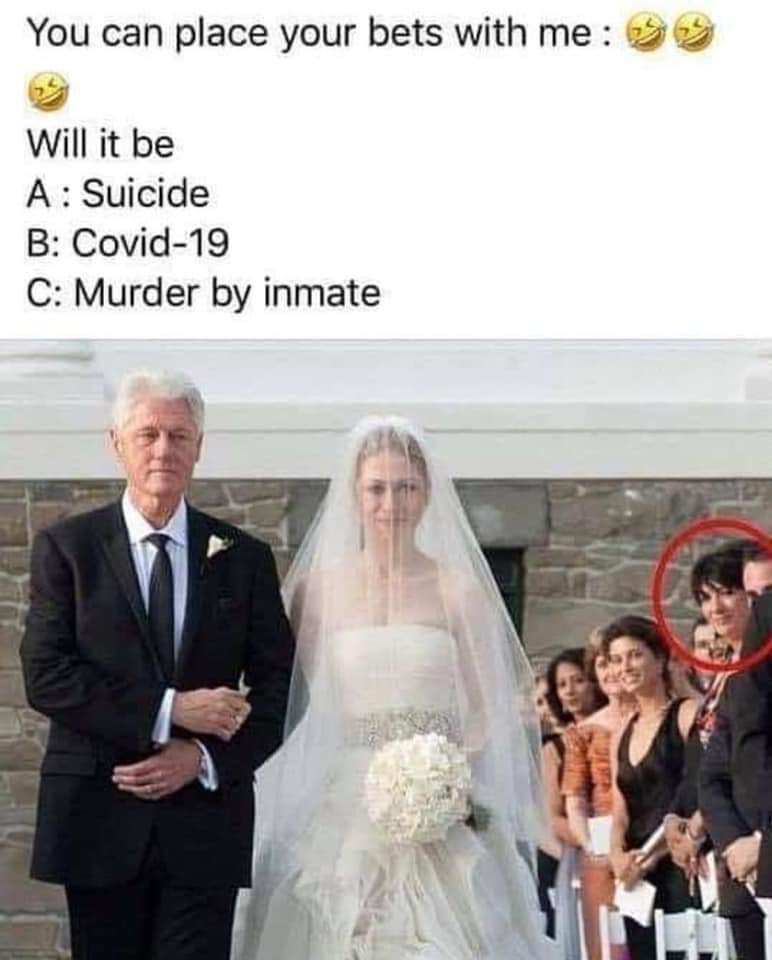}{
\textbf{Ground Truth BKS:}\par%\vspace{0.05em}
1. Bill and Hillary Clinton are married.\par%\vspace{0.05em}
2. People generally invite people they are associated with to their wedding.\par%\vspace{0.05em}
3. Ghislaine Maxwell was a known associate of Jeffrey Epstein.\par%\vspace{0.05em}
4. Jeffrey Epstein allegedly killed himself in prison.\par%\vspace{0.05em}
5. There is a conspiracy theory that he was murdered.\par%\vspace{0.05em}
6. Murders are often covered up by claiming the deceased died from a natural cause. \vspace{0.3cm}
}
&
\contextblock{
\textbf{[Q1]:} Who is Ghislaine Maxwell and what is her connection to Bill Clinton and Chelsea Clinton?\par\vspace{0.05em}
\textbf{[Q2]:} What is the significance of the wedding photograph featuring Bill Clinton, Chelsea Clinton, and Ghislaine Maxwell?\par\vspace{0.05em}
\textbf{[Q3]:} What does the phrase 'Will it be A. suicide B. COVID-19 C. Murder by inmate' refer to in the context of Ghislaine Maxwell?\par\vspace{0.05em}

\ctxcellsep
\textbf{[Our QA-pairs Converted BKS]:} \par\vspace{0.05em}
1. \textbf{Ghislaine Maxwell} is a former girlfriend and accomplice of \textbf{Jeffrey Epstein} who has claimed to have a friendship with Bill Clinton and attended Chelsea Clinton's wedding as a guest.\par\vspace{0.05em}

2. The wedding photograph featuring Bill Clinton, Chelsea Clinton, and Ghislaine Maxwell is significant \textbf{because it shows Maxwell in attendance}, sparking scrutiny and speculation about the relationship between the Clintons and Maxwell.\par\vspace{0.05em}

3. The phrase \emph{``Will it be A. suicide B. COVID-19 C. Murder by inmate''} refers to \textbf{speculation about the cause of death of Jeffrey Epstein,} not Ghislaine Maxwell; it reflects conspiracy theories surrounding Epstein's death in prison, with some suggesting murder rather than suicide.
}
&
\contextblock{
\textbf{[Zero-shot Generated BKS]:}\par\vspace{0.05em}
1. \textbf{Bill Clinton}, the former President of the United States, was involved in a \textbf{scandal} in the 1990s related to an affair with \textbf{Monica Lewinsky}, a White House intern, which led to his impeachment. This scandal is widely known and often referenced in political and cultural discourse.\par\vspace{0.05em}

2. The woman in the wedding dress is Chelsea Clinton, Bill Clinton's daughter, and the image is from her wedding in 2010. The red circle highlights a woman in the background who resembles \textbf{Monica Lewinsky}, which is the source of the meme's humor and controversy.\par\vspace{0.05em}

3. The meme plays on the public's awareness.
}
\\
\hline

\end{tabular}
%}
\caption{Example background knowledge statements generated by our proposed method and the zero-shot baseline. Important information is highlighted in \textbf{bold}.}
\label{tab:bks}
\end{table*}

\subsection{Case Study} \label{ss:case_study}
To explicitly demystify why open-world knowledge decomposition outpaces direct generation, Table~\ref{tab:bks} presents a qualitative comparison between our framework and the zero-shot baseline on a sample from the KYM dataset.~\footnote{The purpose of this case study is not to replace the quantitative evaluation, but to illustrate how question-guided knowledge acquisition changes the type of background knowledge produced compared with direct zero-shot generation.}

The meme features an image of Chelsea Clinton's wedding with a red circle highlighting an unexpected guest, captioned with morbid options regarding an impending death. The zero-shot baseline immediately falls into a \emph{semantic anchoring trap}. Because the parametric weights of the model strongly couple "Bill Clinton" and "scandal" with the 1990s White House accounts, the model misidentifies the circled figure as Monica Lewinsky and hallucinates that the meme's humor is derived from that historic affair. Consequently, the baseline completely misses the actual temporal target, yielding a totally broken context statement.

% The converted background knowledge is not perfect, as some details are not fully supported by the reference annotations. However, it still recovers several key reference-related facts, especially Maxwell's connection to Epstein and speculation about Epstein's death.
Our proposed framework systematically evades this failure mode. By refusing to directly guess the contextual premise, the model instead flags the visual anomaly as an explicit knowledge gap, generating targeted queries: \emph{"Who is the woman... what is the significance of the wedding photograph?"} The resulting web-grounded synthesis accurately surfaces \textit{Ghislaine Maxwell}, her real-world association with Jeffrey Epstein, and the underlying conspiracy theories surrounding mysterious inmate deaths.
% \crcomment{We may need some state-of-the-art models' performance on meme detection as well in Table~\ref{tab:detect_results}}

% \begin{table}[t]
% \centering
% \scriptsize
% \setlength{\tabcolsep}{3pt}
% \begin{tabular}{
% p{1.55cm}
% p{1.55cm}
% |
% >{\centering\arraybackslash}p{1.6cm}
% >{\centering\arraybackslash}p{1.6cm}
% }
% \toprule
% \multicolumn{2}{c|}{\textbf{Model}}
% & \multicolumn{2}{c}{\textbf{Offensiveness Detection on KYM}} \\
% \cmidrule(lr){1-2}
% \cmidrule(lr){3-4}

% \textbf{BKS} & \textbf{Detection}
% & \textbf{Acc.} & \textbf{F1} \\
% \midrule

% \multicolumn{4}{c}{\textbf{Zero-shot without BKS}}\\
% \midrule
% -- & LLaVA-1.5-7B
% & 0.45 & 0.31 \\

% -- & Qwen-VL-8B
% & 0.48 & 0.37 \\

% -- & Gemma3-12B
% & 0.57 & 0.56 \\

% \midrule
% \multicolumn{4}{c}{\textbf{With Zero-shot Generated BKS}}\\
% \midrule
% Qwen-VL-32B & LLaVA-1.5-7B
% & 0.62 & 0.52 \\

% Qwen-VL-32B & Qwen-VL-8B
% & 0.67 & 0.56\\

% Qwen-VL-32B & Gemma3-12B
% & 0.79 & 0.75 \\

% \midrule
% \multicolumn{4}{c}{\textbf{Proposed}} \\
% \midrule
% Qwen-VL-32B & LLaVA-1.5-7B
% & 0.65 & 0.59 \\

% Qwen-VL-32B & Qwen-VL-8B
% & 0.71 & 0.57 \\

% Qwen-VL-32B & Gemma3-12B
% & 0.82 & 0.78 \\

% \bottomrule
% \end{tabular}
% \caption{Experimental results on the KYM offensiveness detection task. The BKS model denotes the model used to generate background knowledge statements; zero-shot detection does not use BKS and is therefore marked with ``--''.}
% \label{tab:kym_offensiveness_results}
% \end{table}

\section{Conclusion}

Our work highlights that meme understanding is not only multimodal reasoning problems, but also open-world knowledge acquisition problems, since many memes depend on recent events, evolving cultural references, and implicit social associations that may be absent from both the meme content and a model's parametric memory. Our results show that explicitly acquiring and grounding missing background knowledge improves meme understanding by recovering more reference-aligned contextual knowledge, and improves meme detection by providing stronger evidence for downstream prediction. This suggests that robust meme analysis requires models to recognize what they do not know, seek external evidence, and use the acquired knowledge to support interpretable and reliable decisions. Future work can extend this direction by evaluating larger and more diverse multilingual meme collections, improving retrieval and evidence verification, reducing inference cost, and incorporating uncertainty estimation or human feedback to build more adaptive, reliable, and culturally aware meme understanding and detection systems.

\section{Limitations}
\noindent\textbf{Dependence on External Evidence. }Our framework relies on the quality of retrieved evidence. While inference-time retrieval helps address dynamic and emerging memes, web and reverse image search may return incomplete, noisy, outdated, or biased results. Errors from captions, entity recognition, or question generation may further propagate into misleading background knowledge.

\noindent\textbf{Limited Evaluation Scope. }
Our experiments mainly focus on English-language memes and social-abuse-related detection tasks, with a relatively small curated KYM dataset. Further validation is needed for multilingual, culturally localized, and low-resource meme communities. LLM-based evidence evaluation may also introduce judgment bias.

\noindent\textbf{Retrieval Cost and Adaptive Planning. }The framework always retrieves from the external web, which increases inference latency and search cost even when parametric knowledge or static knowledge bases may suffice. Future work could develop agentic pipelines that dynamically decide whether to use parametric knowledge, curated knowledge bases, or open-web search for each meme.

\noindent\textbf{Potential Risk.} Because our framework explicitly retrieves and synthesizes background knowledge, it may surface sensitive attributes, or politically and culturally biased associations that are only indirectly implied by a meme. Such knowledge can improve interpretation, but it may also amplify harmful framing or expose sensitive information if not carefully controlled. Future work should incorporate bias-aware filtering, sensitive-content safeguards, and uncertainty signals when using retrieved background knowledge for meme understanding.

\bibliography{custom.bib}

\newpage

\appendix

\addcontentsline{toc}{section}{Appendix}
\startcontents[appendix]

\section*{Contents of Appendix}
\printcontents[appendix]{l}{1}{}

% \section{Creation of KYM}
% \label{app:kym}

\section{Human Alignment of Evaluation Metrics}

\begin{figure*}[t]
    \centering
    \includegraphics[width=0.95\textwidth, trim={0cm 0cm 0cm 0cm}, clip]{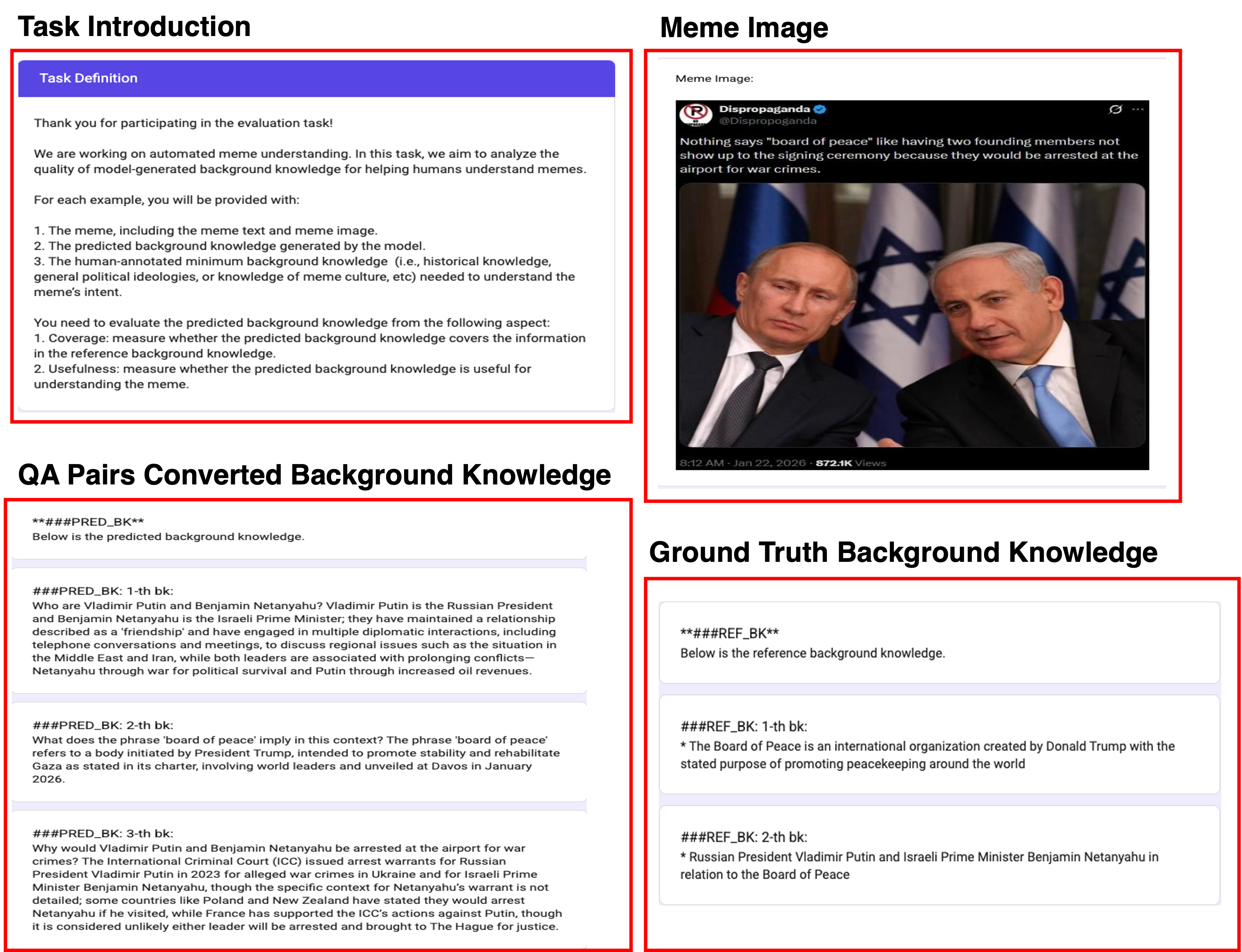}
    \caption{Instructions for Alignment Check on Evidence Evaluation}
    \label{fig:align-human}
\end{figure*}

\subsection{Alignment Check on Evidence Evaluation}
\label{app:align-evidence}
\paragraph{Human Alignment Check.}
To further ensure alignment with human judgment in our setting, we conducted a small-scale human evaluation to validate our reference-based evaluation of generated background knowledge. Specifically, we randomly selected 10 memes from each of the three meme understanding datasets, including our curated KYM dataset, MemeIntent~\cite{park2024memeintent}, and MemeInterpret~\cite{park2025memeinterpret}. The alignment checks were done by four university students and two young professionals based in English speaking countries.

As shown in Figure~\ref{fig:align-human}, for each example, annotators were presented with the meme image, meme text, model-generated background knowledge, and the corresponding human-annotated reference background knowledge. They were asked to compare the generated background knowledge against the reference background knowledge based on two criteria: \textbf{Coverage}, which measures how well the generated background knowledge captures the information contained in the reference background knowledge, and \textbf{Usefulness}, which measures whether the generated background knowledge helps users understand the intended meaning of the meme. As shown in Table~\ref{tab:human_alignment}, the automatic recall-based evaluation achieves Spearman\cite{spearman1961proof} correlations of $\rho_{\mathrm{KYM}}=\text{0.78}$, $\rho_{\mathrm{MemeIntent}}=\text{0.74}$, and $\rho_{\mathrm{MemeInterpret}}=\text{0.77}$, and Pearson\cite{pearson1896vii} correlations of $r_{\mathrm{KYM}}=\text{0.76}$, $r_{\mathrm{MemeIntent}}=\text{0.73}$, and $r_{\mathrm{MemeInterpret}}=\text{0.75}$. These results indicate a
strong alignment between human judgments and the reference-based evaluation scores on the evidence.

\begin{table}[t]
\centering
\resizebox{\columnwidth}{!}{%
\begin{tabular}{lccc}
\toprule
Metric & KYM & MemeIntent & MemeInterpret \\
\midrule
Spearman $(\rho)$ & 0.78 & 0.74 & 0.77 \\
Pearson $(r)$ &0.76   & 0.73 & 0.75 \\
\bottomrule
\end{tabular}
}
\caption{Correlation between human evaluation and automatic recall-based evaluation of generated background knowledge. We report Spearman's rank correlation coefficient $(\rho)$ and Pearson's correlation coefficient $(r)$ for KYM, MemeIntent, and MemeInterpret.}
\label{tab:human_alignment}
\end{table}

\subsection{Alignment Check on Decision Support For Detection Task}
\label{app:align-detect}
To further assess alignment with human judgment, we conduct a \emph{Decision Support} evaluation on MemeInterpret~\cite{park2025memeinterpret}, examining whether the predicted background knowledge provides sufficient information to justify the annotated detection label. Specifically, we treat the detection outputs produced from the predicted evidence, as described in Appendix~\ref{app:align-evidence}, as model predictions, and the raters rate how well the predicted background knowledge supports the given classification label. With a mean Decision Support rating of \textbf{4.28} out of 5.0 (with a standard deviation of $\pm 0.64$), this strong average confirms that our query-driven open-world evidence discovery does not merely collect generic, superficially fluent paragraphs; rather, it successfully isolates and delivers the exact high-fidelity empirical facts required to systematically support and justify downstream classifications.

% \section{Additional Case Studies}
% \label{app:error_analysis}
% Figure~\ref{fig:erroranalysis_understanding}
% \begin{figure*}[t]
%     \centering
%     \includegraphics[width=0.95\textwidth, trim={0cm 0cm 0cm 0cm}, clip]{latex/pic/error_understanding.png}
%     \caption{Examples of zero-recall cases identified by Ev2R evaluation on the KYM, MemeIntent, and MemeInterpret datasets.}
%     \label{fig:erroranalysis_understanding}
% \end{figure*}

\section{Details of Experiments}
\subsection{Computation Resources.}
% \paragraph{Hyper-parameters and Implementation.} In the main experiments, we set.
All experiments are conducted on four NVIDIA L40S GPUs, each with 48~GB of dedicated GPU memory.

\subsection{Experiment Environment and Packages}
In this section, we introduce the experiment environment and packages in use. All experiments are implemented in PyTorch 2.10.0+cu128 with CUDA 12.8. For meme understanding, which involves caption generation, question generation, answer generation, and background knowledge statement generation, we use open-source models loaded through the HuggingFace ecosystem, including \emph{Qwen/Qwen3-VL-32B-Instruct} and \emph{google/gemma-3-27b-it}. For meme detection, we use \emph{llava-hf/llava-1.5-7b-hf}, \emph{Qwen/Qwen3-VL-8B-Instruct}, and \emph{google/gemma-3-12b-it}. The version of \emph{huggingface\_hub} used in our experiments is 1.12.0. For proprietary-model experiments, we access Gemini through the API and use \emph{gemini-3.1-flash-lite}.

\section{Prompts in Use}
In this section, we provide the exact prompts in use for our proposed pipeline and baselines.

\subsection{Prompts for Caption Generation}
\label{app:prompt-caption}
\begin{lstlisting}
You are an Image Captioner for memes. You are given a meme image, its embedded text, and RIS visual hints. Your task is to generate a literal description of what is visibly shown. Use RIS only to help identify ambiguous visible words, people, symbols, or known scenes.

### INPUTS 
Meme text:{text}
RIS visual hints: {ris_summary}

STRICT OUTPUT:
Return JSON ONLY in this format:
{{
  "image_caption": "...",
  "recognized_people": ["..."]
}}

Please generate captions:
\end{lstlisting}

\subsection{Prompts for Question Generation}
\label{app:prompt-question}
\begin{lstlisting}
You are a Knowledge Gap Analyst for meme understanding. You are given a meme image, its embedded text, and Recognized people list (may be empty). Your task is to generate a a small set of useful background knowledge questions that a curious human viewer might ask to better understand the meme.

Please note: The goal is not to explain the meme directly or to fill a fixed number of slots, but to ask the most useful questions that a person would likely search for in order to understand the meme.

### INPUTS 
Meme text: {text}
Recognized people: {recognized_people}
Generated image caption: {gen_caption}

### OUTPUT (JSON ONLY)
{{
  "question_types_used": ["entity", "phrase", "association"],
  "questions": ["...", "...", "..."]
}}

Please generate your question:
\end{lstlisting}

\subsection{Prompts for Answer Generation}
\label{app:prompt-answer}
\begin{lstlisting}
You are answering a question using only the provided retrieved documents. The retrieved documents were specifically retrieved for this question. Your goal is to produce the strongest evidence-grounded answer possible.

Important:
- Use ONLY the provided retrieved documents.
- Do NOT use outside knowledge, memory, assumptions, or unsupported claims.
- If the documents support a full answer, give a concise full answer.
- If the documents support only part of the question, give the best partial evidence-grounded answer.

### INPUTS 
Question: {question}
Retrieved documents: {evidence_block}

Return JSON ONLY in this format:
{{
  "answer": "...",
  "used_chunk_ids": [1, 2]
}}

Please generate your answer:
\end{lstlisting}

\subsection{Prompts for QA Conversion to Evidence Statement}
\label{app:prompt-QA_convert}
Following \cite{cao2025averimatec,akhtar2026ev2r}, we convert QA pairs to evidence statement, for both evidence evaluation and maintaining the evidence history. 
\begin{lstlisting}
You are a expert writer. Given a question ([QUES]) and its answer [ANS], your goal is to convert the caption and the QA pair into a statement [STAT].
Below are some examples:

[QUES]: What event or incident is referenced by the image of the Statue of Liberty with dark smoke rising behind a city skyline?
[ANS]: The image of the Statue of Liberty with dark smoke rising behind a city skyline refers to the events following the 9/11 terrorist attacks in Lower Manhattan.
[STAT]: The event referenced by the image of the Statue of Liberty with dark smoke rising behind a city skyline is the 9/11 terrorist attacks in Lower Manhattan.

[QUES]: What is the significance of \"MN\" in the context of this meme?
[ANS]: In the context of the meme, \"MN\" refers to Minnesota, as evidenced by references to Minneapolis, Minnesota.
[STAT]: \"MN\" refers to Minnesota, a U.S. state in the context of this meme.

[QUES]: Who is George Floyd, and how is him relevant to Minnesota in this meme?
[ANS]: George Floyd was an African American man. He was killed during an arrest by Minneapolis police in May 2020. His death led to large protests in Minneapolis.
[STAT]: George Floyd  was an African man. He is relevant to Minnesota in the meme because he died during a police arrest in Minneapolis, and the incident became widely associated with protests against police violence.

Please convert the QA pair below into its statement:
[QUES]: {ques}
[ANS]: {ans}
[STAT]:
\end{lstlisting}

\subsection{Prompts for Evidence Evaluation.}
\label{app:prompt-evidence_eval}
As described in Section\ref{experiment}, we compare the textual of retrieved evidence and ground-truth evidence.

\begin{lstlisting}
You will get as input a reference evidence ([REF]) and a predicted evidence ([PRED]). 
The predicted evidence contains generated statements. The reference evidence contains the reference background knowledge statements. Please verify the correctness of the predicted evidence by comparing it to the reference evidence, following these steps: 1. Evaluate each fact in the predicted evidence individually: is the fact supported by the REFERENCE evidence? Do not use additional sources or background knowledge. 2. Evaluate each fact in the reference evidence individually: is the fact supported by the PREDICTED evidence? Do not use additional sources or background knowledge. 3. Finally summarise (1.) how many predicted facts are supported by the reference evidence and explanations ([PRED in REF] and [PRED in REF Exp]), (2.) how many reference facts are supported by the predicted evidence and explanations ([REF in PRED] and [REF in PRED Exp]). Generate the output as shown in the examples below:

[PRED]: 1. Donald Trump is an American politician, media personality, and businessman who is the 47th president of the United States. 2. The phrase "It affects virtually nobody. It's an amazing thing." refers to the coronavirus (COVID-19), as stated by Donald Trump while downplaying the severity of the virus.

[REF]: 1. Trump has been accused of downplaying the severity of COVID. 2. Many people have been negatively affected by COVID.
[PRED in REF]: 1
[PRED in REF Exp]: 1. The first predicted statement provides general biographical information about Donald Trump, which is not directly supported by the reference evidence. 2. The second predicted statement is supported by the reference evidence because the reference states that Trump has been accused of downplaying the severity of COVID, and that many people have been negatively affected by it.
[REF in PRED]: 2
[REF in PRED Exp]: 1. The first reference fact is supported by the second predicted statement, which states that Trump downplayed the severity of COVID. 2. The second reference fact is supported by the second predicted statement, which refers to COVID and implies its serious impact.

[PRED]: 1. Ilhan Omar is an American politician and member of the U.S. House of Representatives. 2. Ilhan Omar's statement "I hate Trump" is being paired with the response "Most Terrorists do" as a rhetorical criticism that associates her statement with extremist views.
[REF]: 1. Ilhan Omar is a U.S. congresswoman from Minnesota. 2. The meme frames Ilhan Omar negatively by linking her anti-Trump statement to terrorism.
[PRED in REF]: 2
[PRED in REF Exp]: 1. The first predicted statement is supported by the first reference fact, since both state that Ilhan Omar is a U.S. politician/congresswoman. 2. The second predicted statement is supported by the second reference fact, since both describe the meme as linking Omar's anti-Trump statement with terrorism in a negative framing.
[REF in PRED]: 2
[REF in PRED Exp]: 1. The first reference fact is supported by the first predicted statement. 2. The second reference fact is supported by the second predicted statement.

[PRED]: 1. The phrase "THE PARTY OF DIVERSITY" refers to the Democratic Party. 2. The labels "GALAXY DUST" and "MOSTLY GAS" are used in the meme to suggest exaggerated or mocking identity labels.
[REF]: 1. The meme critiques the Democratic Party's claim to diversity. 2. The meme uses unusual labels to mock identity-based politics.
[PRED in REF]: 2
[PRED in REF Exp]: 1. The first predicted statement is supported by the first reference fact because both refer to the Democratic Party and its claim to diversity. 2. The second predicted statement is supported by the second reference fact because both describe the unusual labels as part of the meme's mockery.
[REF in PRED]: 2
[REF in PRED Exp]: 1. The first reference fact is supported by the first predicted statement. 2. The second reference fact is supported by the second predicted statement.

Return the output in the exact format as specified in the examples, do not generate any additional output:
[PRED]: [PRED_EVID]
[REF]: [REF_EVID]

\end{lstlisting}

\subsection{Prompts for Zero-shot Background Knowledge Generation}
\label{app:prompt-zsbks}
We adopt the zero-shot background knowledge generation prompt from \cite{park2025memeinterpret}.
\begin{lstlisting}
You will be provided with a meme and its embedded text. Your task is to infer the background knowledge that a reader of the meme needs to possess before they can understand the ultimate intent behind the creation or sharing of a meme, as perceived by its audience. Background knowledge is the minimum amount of knowledge that is missing from the meme. It is the knowledge that needs to be combined with visual and textual cues from the meme in  order to munderstand its meaning. Give me background knowledge in the form of a list. For example: '1. Soccer is the sport that  children like a lot. 2. There are two main political parties in the US: Democratic and Republican.' Each background knowledge item must be in one to three sentences.
### Background knowledge:
\end{lstlisting}

\subsection{Prompts for Detection}
\label{app:prompt-detect}
Recall that we have defined the detection task as: 
\[
\mathcal{S} \in 
\left\{
\begin{aligned}
&\text{hatefulness},\ \text{misogyny},\ \text{offensiveness},\\
&\text{sarcasm},\ \text{harmfulness},\ \text{humor}
\end{aligned}
\right\},
\]
Inspired by \cite{lin2024goat} , for each task $\ast \in \mathcal{S}$, we standardize the detection prompt by conditioning on the meme content as follows:
\begin{itemize}
    \item For Vanilla Zero-shot Detection: \\
    \emph{``Given the meme, with the text [$\mathcal{T}$] accompanied by the image [$\mathcal{I}$], is this meme [$\ast$?]''}
    
    \item For Detection with Background Knowledge: \\
    \emph{``Given the meme, with the text [$\mathcal{T}$] accompanied by the image [$\mathcal{I}$], the background knowledge [$\mathcal{B}$],\footnote{For detection with zero-shot generated background knowledge, $\mathcal{B}$ is generated by ~\ref{app:prompt-zsbks}. In our proposed pipeline, $\mathcal{B}$ is the background-knowledge statements converted from the QA pairs.} is this meme [$\ast$?]''}. 
\end{itemize}
Below are the prompts that are used:
    \begin{lstlisting}
Given the meme, with the text [{text}] accompanied by the image [{img_name}], is this meme {task_word}? 
Return exactly one character: 
1 if yes
0 if no
Do not output any other word.
\end{lstlisting}

    \begin{lstlisting}
Given the meme, with the text [{text}] accompanied by the image [{img_name}], the background knowledge of the meme [{background_knowledge}], is this meme {task_word}?
Return exactly one character:
1 if yes
0 if no
Do not output any other word.

\end{lstlisting}

\section{Declaration on AI Assistance}

AI-assisted writing tools were used to polish language, correct grammar, and improve the organization and clarity of the manuscript. The research design, experiments, analysis, and conclusions were conducted and verified by the authors. The authors reviewed all AI-assisted edits and take full responsibility for the final content.

\end{document}